\documentclass[journal]{IEEEtran}

\usepackage{cite}
\usepackage{algorithm}
\usepackage{algorithmic}
\usepackage{amsmath}
\usepackage{graphicx}
\usepackage{subcaption}
\usepackage{multirow}

\usepackage{xcolor}

\DeclareMathOperator*{\argmax}{argmax}

\title{Active Learning of Causal Structures with Deep Reinforcement Learning}
\author{Amir Amirinezhad,~Saber Salehkaleybar, and~Matin Hashemi
	
	\thanks{
		Authors are with Learning and Intelligent Systems Laboratory, Department
		of Electrical Engineering, Sharif University of Technology, Tehran, 
		Iran. \protect \\
		E-mails: amirinezhad.amir@ee.sharif.edu, saleh@sharif.edu (corresponding author), matin@sharif.edu.  Webpage: http://lis.ee.sharif.edu 
	}
}

\newtheorem{example}{Example}

\begin{document}

\maketitle

\begin{abstract}
We study the problem of experiment design to learn causal structures from interventional data. We consider an active learning setting in which the experimenter decides to intervene on one of the variables in the system in each step and uses the results of the intervention to recover further causal relationships among the variables. The goal is to fully identify the causal structures with minimum number of interventions. We present the first deep reinforcement learning based solution for the problem of experiment design. In the proposed method, we embed input graphs to vectors using a graph neural network and feed them to another neural network which outputs a variable for performing intervention in each step. Both networks are trained jointly via a Q-iteration algorithm. Experimental results show that the proposed method achieves competitive performance in recovering causal structures with respect to previous works, while significantly reducing execution time in dense graphs.
\end{abstract}

\begin{IEEEkeywords}
	Casual structure learning, Experiment design, Active learning, Deep reinforcement learning
\end{IEEEkeywords}

\section{Introduction}
\label{sec:introduction}

\IEEEPARstart{R}{ecovering} causal relations among a set of variables in various natural or social phenomena  is one of the primary goals in artificial intelligence. For instance, one might be interested in estimating the effect of education on salaries, smoking on lung cancer, or activation of genes on a phenotype. If we have only access to observational data (in contrast to possibility of intervening in the system), a part of causal relationship can be identified in most cases. As a result, the investigator is mostly left with some unresolved causal relations. However, if we could perform experiments sufficiently in the system, all the causal relationships can be recovered. Unfortunately, in many applications, it might be too costly to intervene in the system \cite{hoyer2008estimation}. Thus, it is desirable to design optimal experiment, i.e., a set of interventions with minimum size that results in full identification of causal relationships.

Directed acyclic graphs (DAGs) are commonly used to present causal structures where each node is a random variable and there is a directed edge from a variable $X$ to variable $Y$ if $X$ is a direct cause of $Y$. From the observational data, the underlying causal DAG can be identified up to a Markov Equivalence Class (MEC), which is the set of DAGs representing the same set of conditional independencies among the variables \cite{pearl2009causality}. Several methods \cite{spirtes2000causation,pearl2009causality,chickering2002optimal} in the literature have been proposed to learn MEC from purely observational data. However, in order to uniquely identify the causal structure, it is necessary to intervene in the system if there is no prior assumption on the data generation mechanism. In most applications, performing experiments are too costly or even infeasible. Thus, we need to fully identify the true causal structure with minimum number of interventions. 

Eberhardt \cite{eberhardt2007causation} presented the worst-case bounds on the number of required experiments for full identification where the number of intervened variables could be as large as half of the size of graph. He and Geng \cite{he2008active} considered the problem of experiment design in two passive and active settings. In the passive setting, all the experiments are designed before doing interventions in the system. Afterwards, the results of interventions are aggregated in order to recover the causal structure. While, in the active setting, we sequentially perform interventions and the results of previous interventions help in designing latter ones. He and Geng \cite{he2008active} assumed that orientations of edges incident to an intervened variable can be revealed by performing perfect randomized interventions. In the passive setting, they enumerated all possible DAGs in the MEC obtained from observational data for a given experiment and checked whether it can fully identify the causal structure for any DAG in the equivalence class. From experiments with such desirable property, they returned the one with minimum number of interventions. Unfortunately, the proposed method might be too computationally intensive due to possibly large number of DAGs in an MEC. In the active setting, they also proposed a heuristic algorithm that decides which variable to be intervened on in the next step based on Shannon's entropy metric. 
Hauser and Buhlmann \cite{hauser2012characterization} studied the problem of experiment design in the active setting where the experimenter is allowed to do intervention on a single variable in each step. They proposed an optimal algorithm for the single step case and used it as a heuristic for selecting variables for the case of multiple steps. Shanmugam et al. \cite{shanmugam2015learning} considered the problem of causal structure learning by performing experiments with bounded number of interventions in each experiment. They derived lower bounds on the number of experiments for full identification of causal graphs in both passive and active settings using the theoretical results in separating systems. Kocaoglu et al. \cite{kocaoglu2017cost} considered the problem of experiment design with no constraints on the number of interventions in each experiment and proposed an algorithm for the case that there is a specific cost for intervening on each variable. Ghassami et al. \cite{ghassami2018budgeted} proposed an approximate algorithm in the passive setting which maximizes the average number of oriented edges for a fixed budget of interventions. Later, this approach has been accelerated using clique trees \cite{ghassami2019counting} and efficient iteration on chain components \cite{icml2020_1030}. Recently, Agrawal et al. \cite{agrawal2019abcd} proposed a Bayesian experiment design algorithm in the active setting where the expected value of a utility function is maximized in each step according to the current belief and they proposed a tractable solution with an approximation guarantee based on sub-modular functions.

Several works utilized reinforcement learning in order to train an agent for solving NP-Hard problems on graphs \cite{khalil2017learning,abe2019solving,kool2018attention,nazari2018reinforcement,chen2019learning}. For instance, Dai et al. \cite{khalil2017learning} embedded graphs to a vector by a graph neural network and fed it to another neural network in order to form a solution for difficult graph problems such as minimum vertex cover or traveling salesman problem. 
Abe et al. \cite{abe2019solving} combined Monte-Carlo tree search and graph isomorphism networks to tackle NP-hard problems. In \cite{kool2018attention,nazari2018reinforcement,chen2019learning}, several approaches based on reinforcement learning have been proposed to solve the vehicle routing problem. 
Experimental results showed these algorithms can outperform traditional heuristic methods in terms of the quality of solution. Besides, the running time of solving a new instance of the problem is significantly reduced after the training phase, while most heuristic methods become computationally intensive in large graphs. 

Previous works on experiment design mainly focused on developing heuristic metrics to decide which variables should be intervened on and most of these metrics are related to graph properties of MEC. In this paper, we consider the problem of experiment design in the active setting. Unlike previous works, our goal is to train an agent by utilizing reinforcement learning algorithm in order to decide which variable is suitable for intervention in each step. In particular, we embed input graphs to vectors using a graph neural network and feed them to another neural network which outputs a variable for performing intervention in the next step. We jointly train both neural networks via a Q-iteration algorithm. Our experiments on synthetic and real graphs show that the proposed method achieves competitive performance in recovering causal structures while it reduces running times by a factor up to $757$.

The structure of the paper is as follows: In Section 2, we review some preliminaries on causal structures and define the problem of experiment design in the active learning setting. In Section 3, we present our proposed method and describe training algorithm. In Section 4, we provide experimental results and compare our method with previous works. We conclude the paper in Section 5.

\section{Problem definition}
\subsection{Preliminaries}
A graph $G$ is represented by a pair $G=(V(G),E(G))$ where $V(G)$ is the set of vertices and $E(G)$ is the set of edges. There exists an undirected edge between two vertices $X$ and $Y$ if $(X,Y)\in E(G)$ and $(Y,X)\in E(G)$. Moreover, there is a directed edge from vertex $X$ to vertex $Y$ if $(X,Y)\in E(G)$ while $(Y,X)\not \in E(G)$. We denote the directed edge from $X$ to $Y$ and undirected edge between $X$ and $Y$ by $X\rightarrow Y$ and $X-Y$, respectively. If there is a directed edge from $X$ to $Y$, we consider $X$ as a parent of $Y$. Decedents of $X$ are the set of vertices with a directed path from $X$ to each variable in the set. We say a graph $G$ is a directed graph if all its edges are directed. A sequence $(X_1,\cdots,X_k)$ is a partially directed path in a graph $G$ if either $X_i\rightarrow X_{i+1}$ or $X_i-X_{i+1}$  for all $i=1,\cdots,k-1$. A partially directed cycle is a partially directed path where the first and last vertices in the path are the same vertex. We say that a graph is a chain graph if it does not contain any partially directed cycle. After removing directed edges of a chain graph, the remaining undirected connected components are called the chain connected components of the graph. 
Furthermore, an undirected graph is a chordal if every cycle of length four or greater has a chord. It can be shown that each chain connected component is chordal.  A v-structure is a sub-graph of $G$ with three vertices $X,Y,$ and $Z$ such that $X\rightarrow Z\leftarrow Y$. Two directed graphs have the same skeleton if they have the same set of vertices and edges regardless of their orientations. 

Let $\mathcal{X}=\{X_1,\cdots,X_n\}$ be a set of random variables. Consider a graph $G$ whose set of vertices is equal to $\mathcal{X}$. A joint distribution $P$ over $\mathcal{X}$ satisfies Markov property with respect to $G$ if any variable of $G$ is independent of its non-descendants given its parents. Under causal sufficiency and faithfulness assumptions \cite{pearl2009causality}, any conditional independence in $P$ can be inferred by Markov property. Furthermore, multiple DAGs may encode a same set of conditional independence assertions. A Markov equivalence class (MEC) is a set of DAGs entailing the same set of conditional independence assertions. The set of all DAGs that are Markov equivalent to some DAG $G$ can be represented by a completed partially DAG (CPDAG) in which there is a directed edge from $X$ to $Y$ if for all DAGs in MEC, $X$ is a parent of $Y$. Otherwise, this edge is represented by an undirected edge. It can be shown that all DAGs in an MEC have the same skeleton and the same set of v-structure \cite{verma1991equivalence}. Moreover, CPDAG can be obtained from the skeleton and the set of v-structures by applying commonly called Meek rules \cite{meek1997graphical} which orient further edges such that no directed cycle or a new v-structure is created. 

\begin{figure}[t!]
	\centering
	\begin{subfigure}[b]{0.45\linewidth}
		\includegraphics[width=\linewidth]{{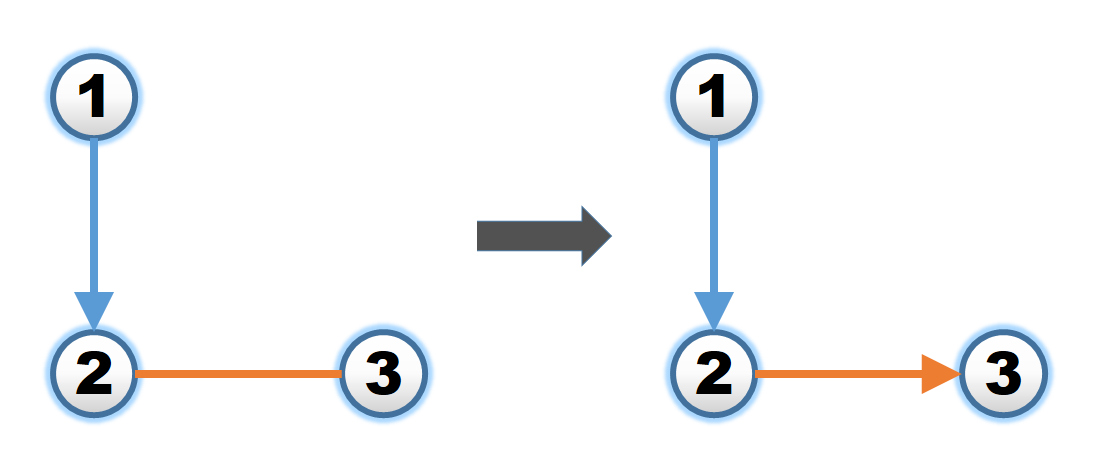}}
		\caption{Rule 1}
		\label{subfig:rule1}
	\end{subfigure}
	\begin{subfigure}[b]{0.45\linewidth}
		\includegraphics[width=\linewidth]{{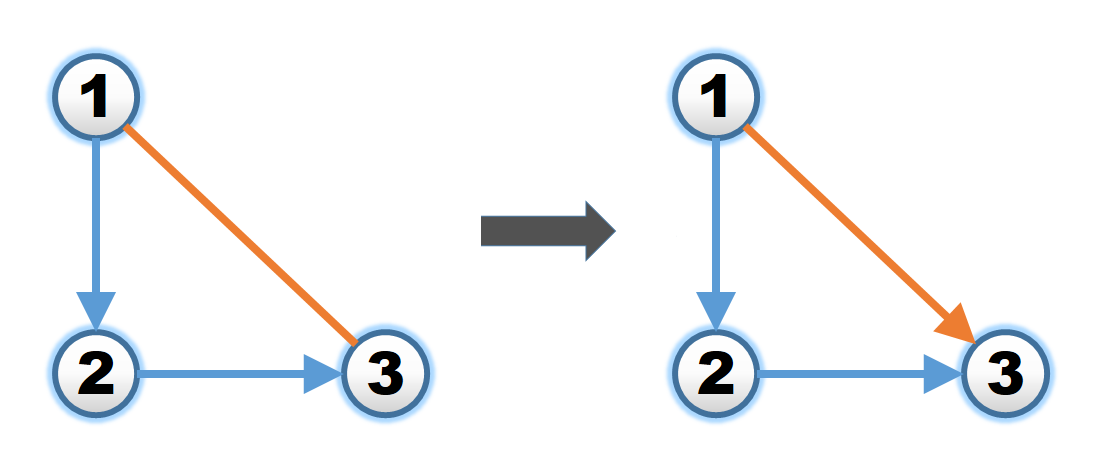}}
		\caption{Rule 2}
		\label{subfig:rule2}
	\end{subfigure}
	\begin{subfigure}[b]{0.45\linewidth}
		\includegraphics[width=\linewidth]{{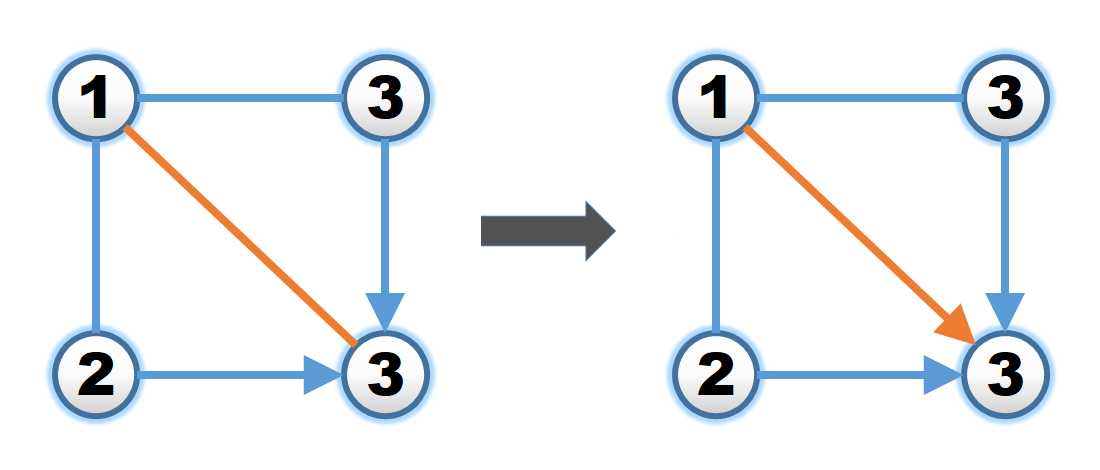}}
		\caption{Rule 3}
		\label{subfig:rule3}
	\end{subfigure}
	\begin{subfigure}[b]{0.45\linewidth}
		\includegraphics[width=\linewidth]{{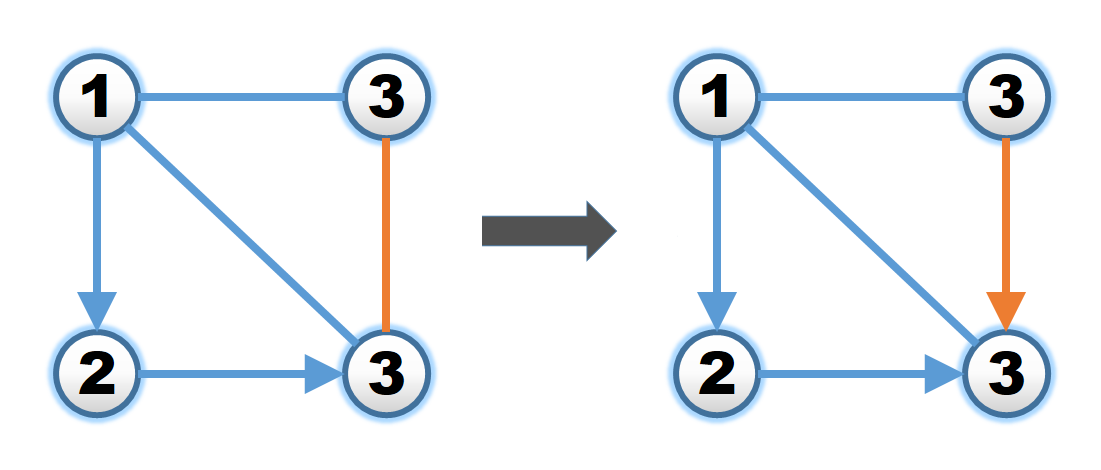}}
		\caption{Rule 4}
		\label{subfig:rule4}
	\end{subfigure}
	\caption{Meek rules\cite{meek1997graphical}: If the graph on the left-hand side of each sub-figure is an induced subgraph of a CPDAG $G$, then we can orient the undirected edge in the orange color as shown in the right-hand side of the sub-figure.  }
	\label{fig:meek_rules}
\end{figure}

Fig.~\ref{fig:meek_rules} presents Meek rules that are used in obtaining a CPDAG. For instance, if there is a subgraph like the one in the left-hand side of Fig.~\ref{fig:meek_rules}a, we can orient the edge between node $2$ and $3$ and obtain the subgraph in the right-hand side of Fig.~\ref{fig:meek_rules}a. Otherwise, we create a new v-structure in the CPDAG. It can be shown that the CPDAG can be obtained from applying these four Meek rules in any order until no more further edges can be oriented \cite{meek1997graphical}.

\begin{figure}[t]
	\centering
	\begin{subfigure}[b]{0.6\linewidth}
		\includegraphics[width=\linewidth]{{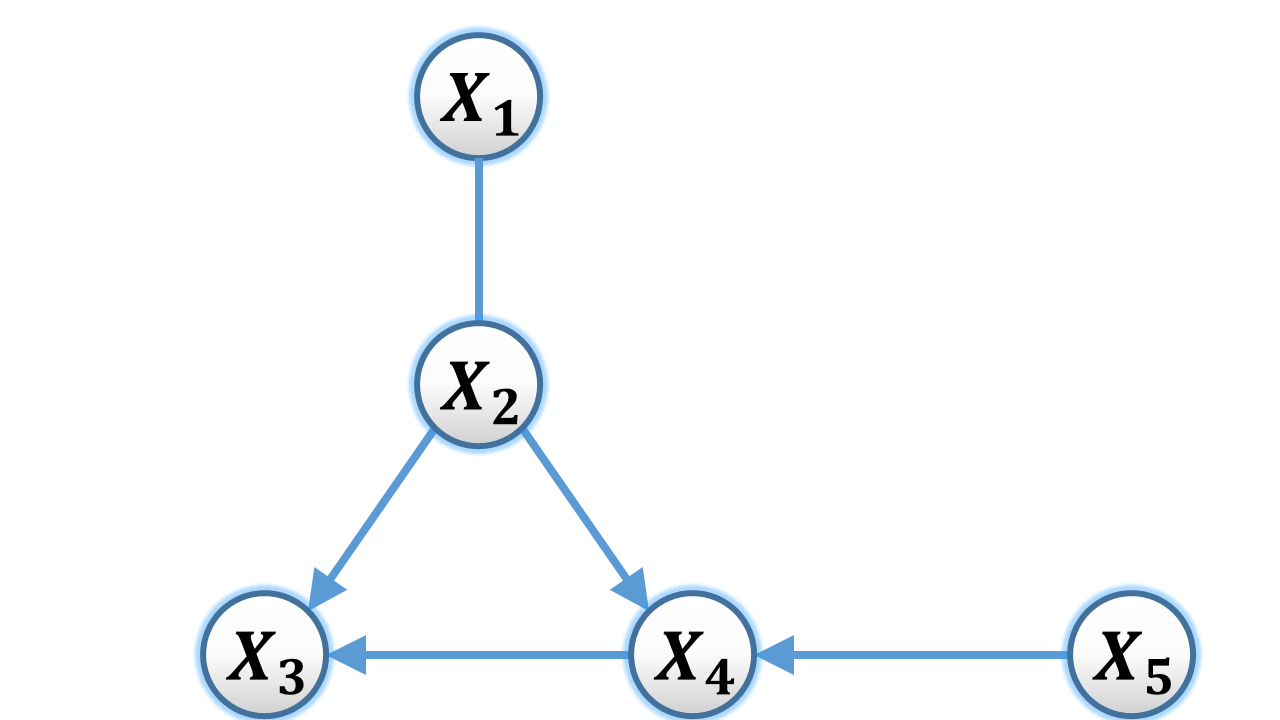}}
		\caption{A CPDAG}
		\label{subfig:cpdag}
	\end{subfigure}

	\begin{subfigure}[b]{0.6\linewidth}
		\includegraphics[width=\linewidth]{{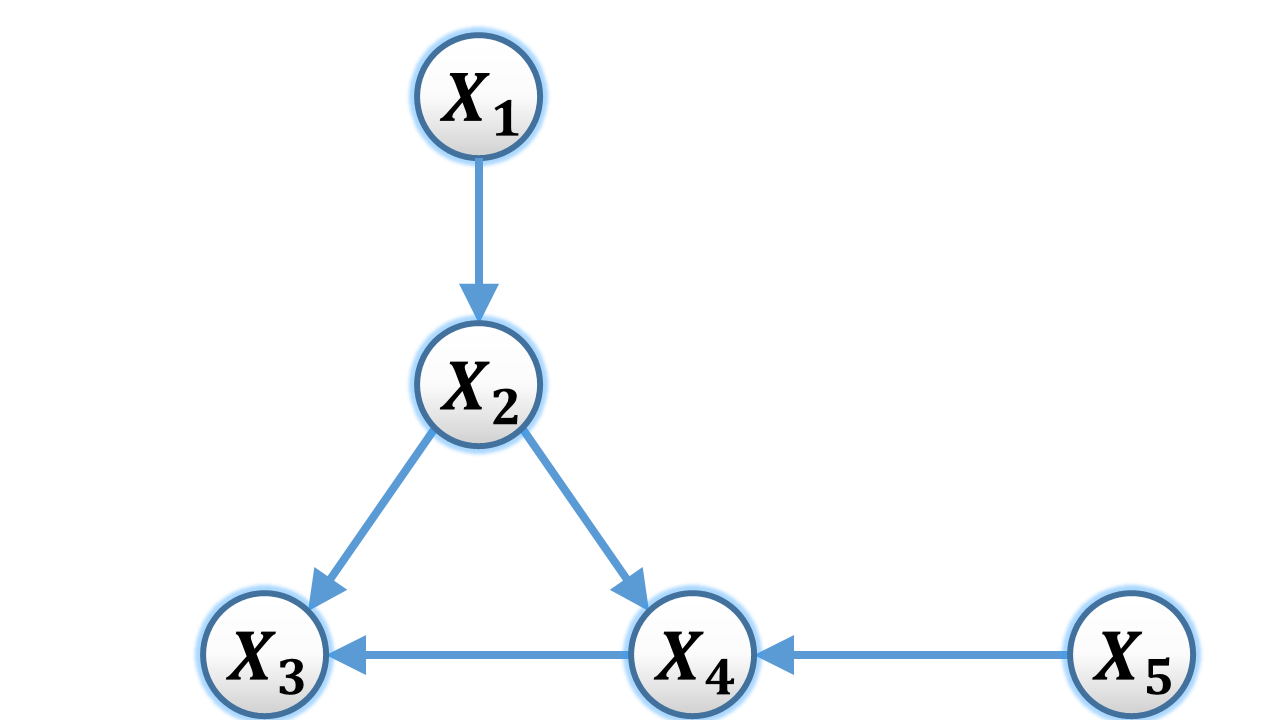}}
		\caption{DAG1}
		\label{subfig:dag1}
	\end{subfigure}
	\begin{subfigure}[b]{0.6\linewidth}
		\includegraphics[width=\linewidth]{{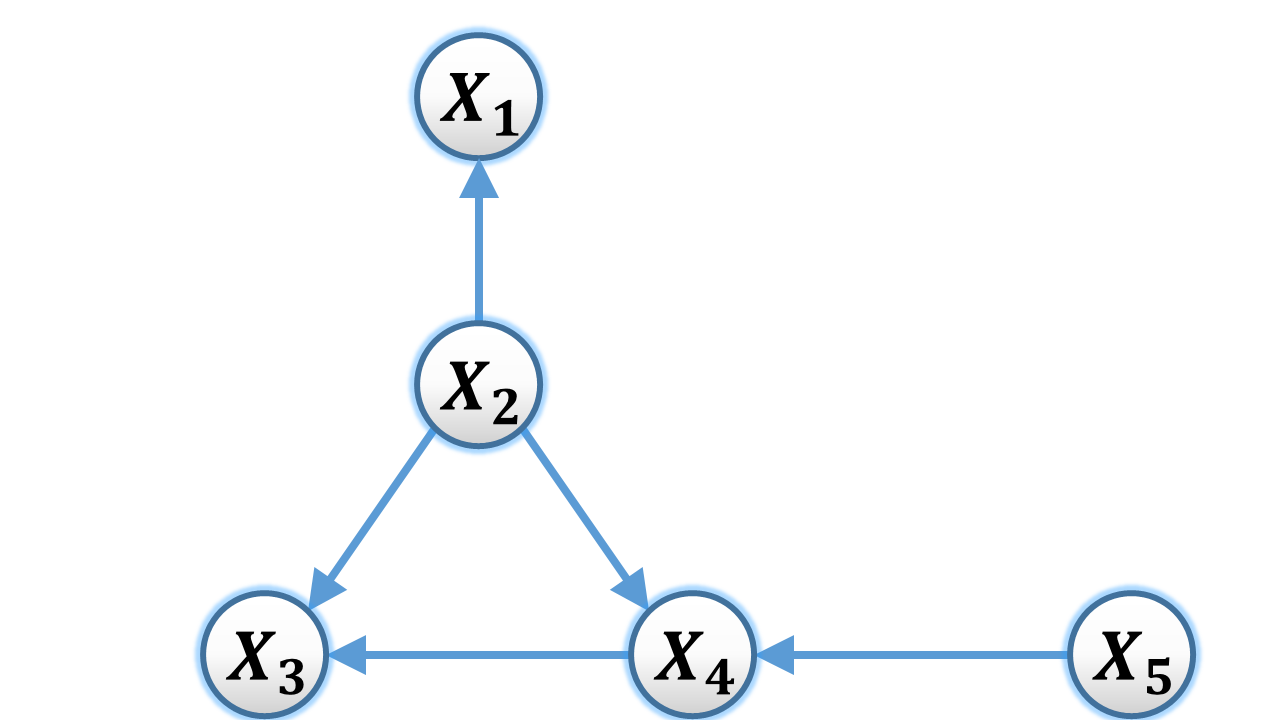}}
		\caption{DAG2}
		\label{subfig:dag2}
	\end{subfigure}
	\caption{(a) An example of a CPDAG with a single v-stucture. Directed graphs in (b) and (c) are two DAGs of the CPDAG in (a). }
	\label{fig:example_cpdag}
\end{figure}

\begin{example}
An example of a CPDAG and the corresponding DAGs are given in \figurename~\ref{fig:example_cpdag}. In the CPDAG shown in \figurename~\ref{subfig:cpdag}, there is only one v-structure $X_2\rightarrow X_4 \leftarrow X_5$. In the DAG in \figurename~\ref{subfig:dag1}, the variable $X_1$ is the root variable (a variable with no incoming edges) while in \figurename~\ref{subfig:dag2}, the root variable is $X_2$. Please note that we can orient the edge between $X_1$ and $X_2$ in either direction since we are not creating any new v-structure or cycle. Moreover, the direction of the edge between $X_3$ and $X_4$ should be from $X_4$ to $X_3$ in both DAGs. Otherwise, it results in a new v-structure. Furthermore, the direction of the edge between $X_2$ and $X_3$ should be from $X_2$ to $X_3$ to avoid creating any cycle. 
\end{example}

\subsection{Problem of Experiment Design}	
			
Let $G^*$ be the underlying causal graph among the variables in $\mathcal{X}$. From the observational data, one can just recover the true causal graph $G^*$ up to a MEC through constraint-based approaches \cite{pearl2009causality, spirtes2000causation} or score-based approaches \cite{meek1997graphical, chickering2002optimal, tian2012bayesian, solus2017consistency}. Thus, the orientations of undirected edges in CPDAG cannot be identified by merely observational data. In order to fully recover the whole causal graph, it is required to perform experiments to orient further edges. In most applications, intervening on variables might be costly or time consuming. Thus, it is vital to recover the causal graph with minimum number of intervention.

In this paper, we consider active learning setting for performing experiments. In particular, in the first step, the CPDAG of true MEC containing the causal graph is obtained from observational data which we denote it by $G_0$. 
 Next, in each step $j$ of active learning, we select a variable $X_{i_j}$ from $G_j$ to be intervened on. By performing perfect randomized experiment, we assume that the orientations of all incident edges to $X_{i_j}$ are identified. By considering the orientations of these edges in $G_j$, we can apply meek rules to recover the orientations of further edges. Let $G'_j$ be the resulted causal graph. We obtain $G_{j+1}$ by removing the oriented edges in $G'_j$. It can be shown that $G_{j+1}$ is a collection of undirected chain components. In the next step, we decide to intervene on one of the variables in the collection of chain components. This procedure continues  until the whole causal graph is recovered. Our main goal is to select the intervened variable $X_{i_j}$ in each step $j$ such that the total number of steps until identifying the true causal graph is minimized.

\begin{figure}[t]
	\centering
	\begin{subfigure}[b]{0.65\linewidth}
		\includegraphics[width=\linewidth]{{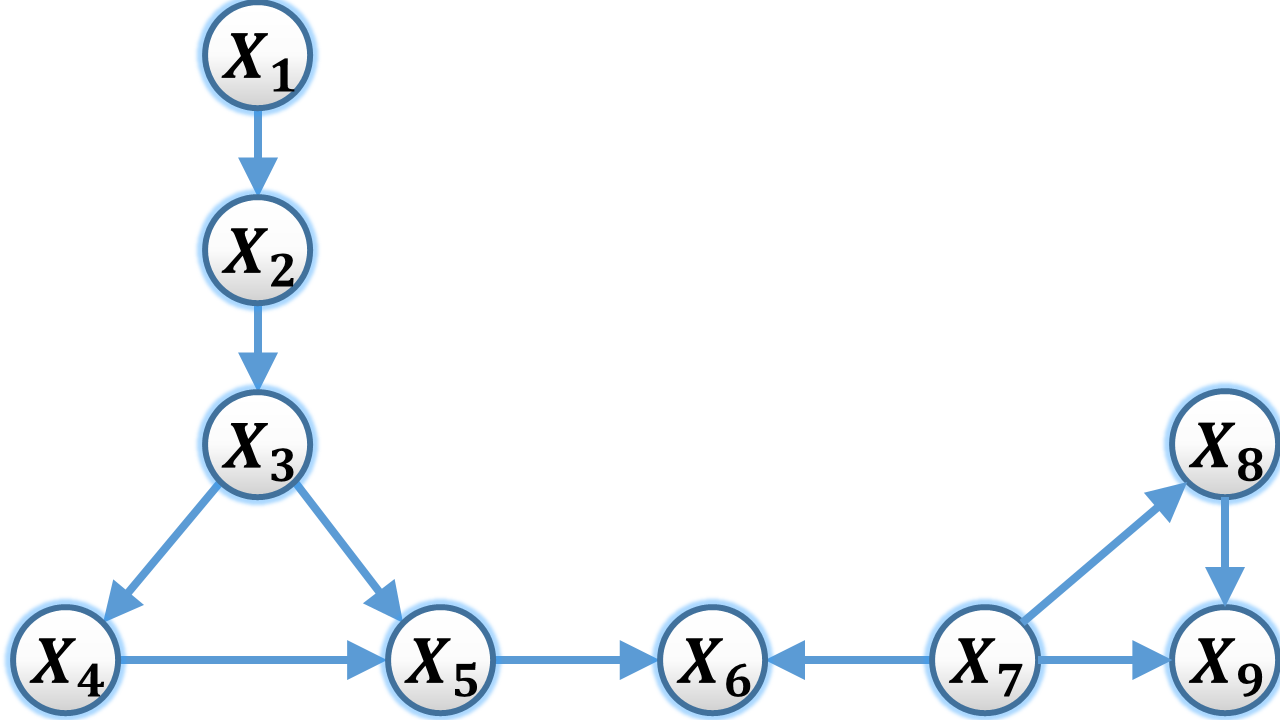}}
		\caption{$G^*$}
	\end{subfigure}
	\begin{subfigure}[b]{0.65\linewidth}
		\includegraphics[width=\linewidth]{{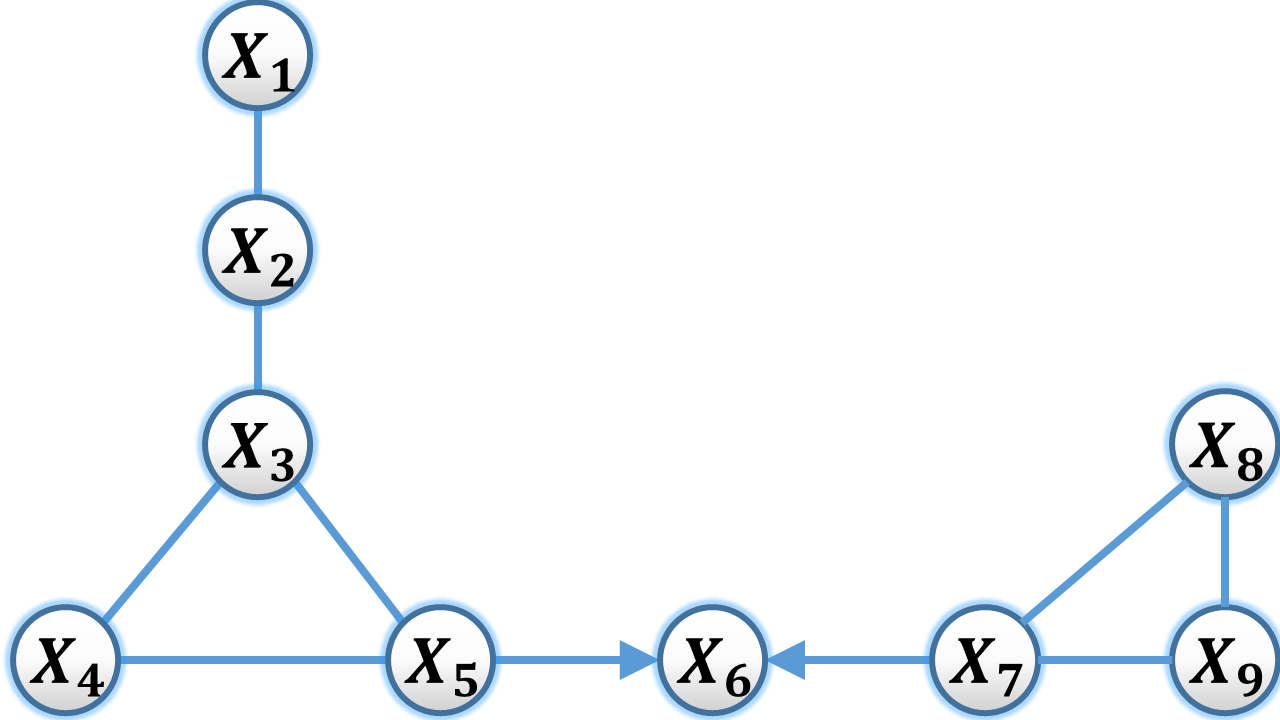}}
	\caption{$G_0$}
	\end{subfigure}
	\caption{An example of $G^*$ and corresponding CPDAG denoted by $G_0$.}
	\label{fig:example}
\end{figure}

\begin{figure}[h!]
	\centering
	\begin{subfigure}[b]{0.6\linewidth}
		\includegraphics[width=\linewidth]{{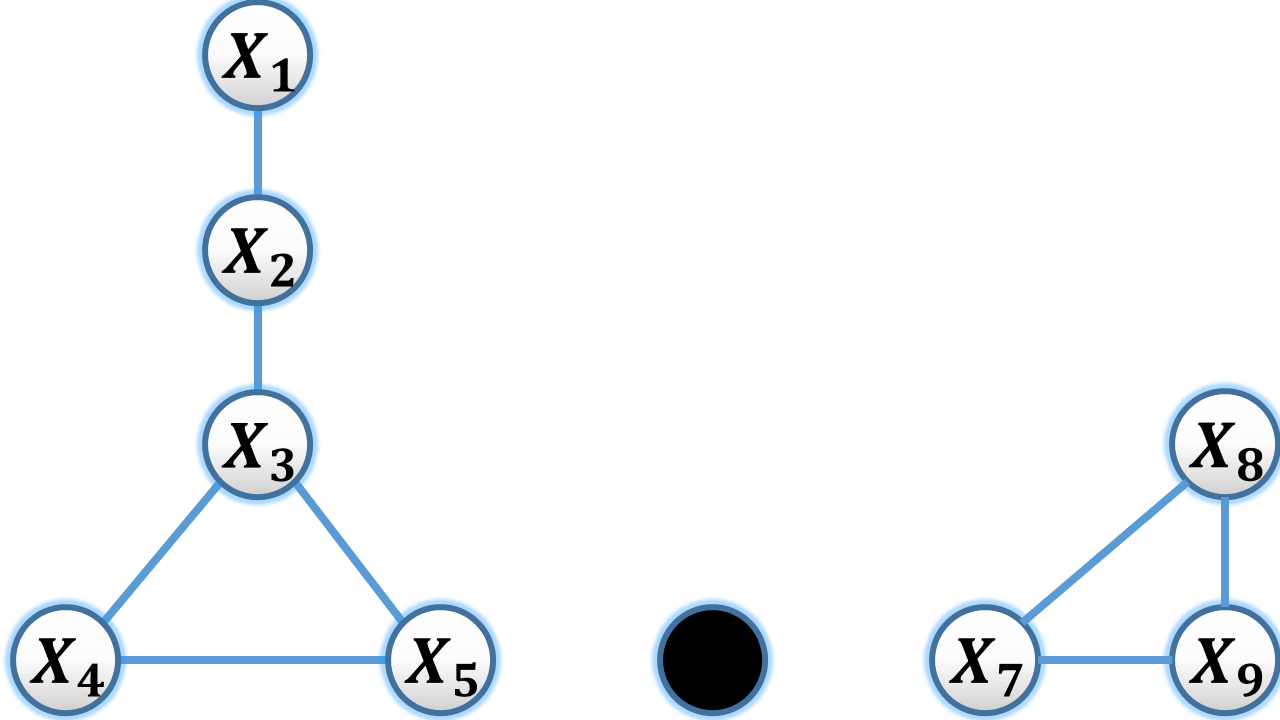}}
		\caption{$G_1$}
	\end{subfigure}
	\begin{subfigure}[b]{0.6\linewidth}
		\includegraphics[width=\linewidth]{{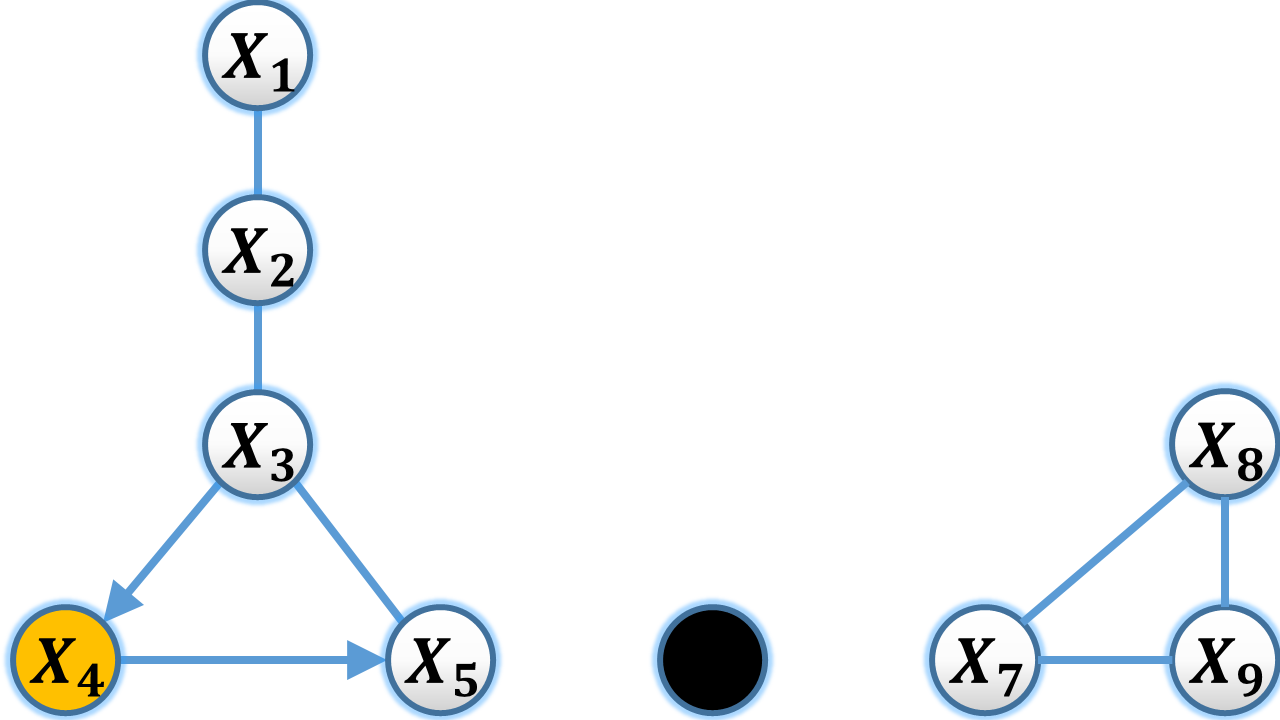}}
		\caption{Intervening on node 4.}
	\end{subfigure}
	\begin{subfigure}[b]{0.6\linewidth}
		\includegraphics[width=\linewidth]{{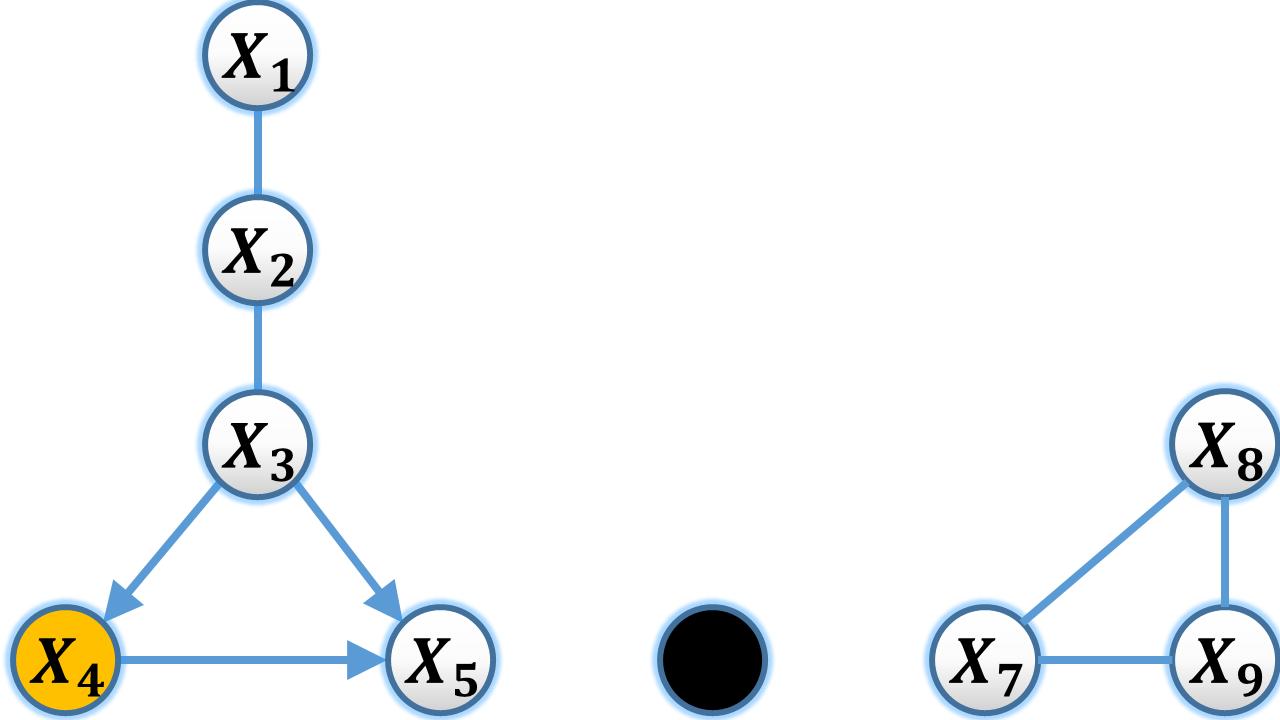}}
		\caption{Applying meek rules.}
	\end{subfigure}
	\begin{subfigure}[b]{0.6\linewidth}
		\includegraphics[width=\linewidth]{{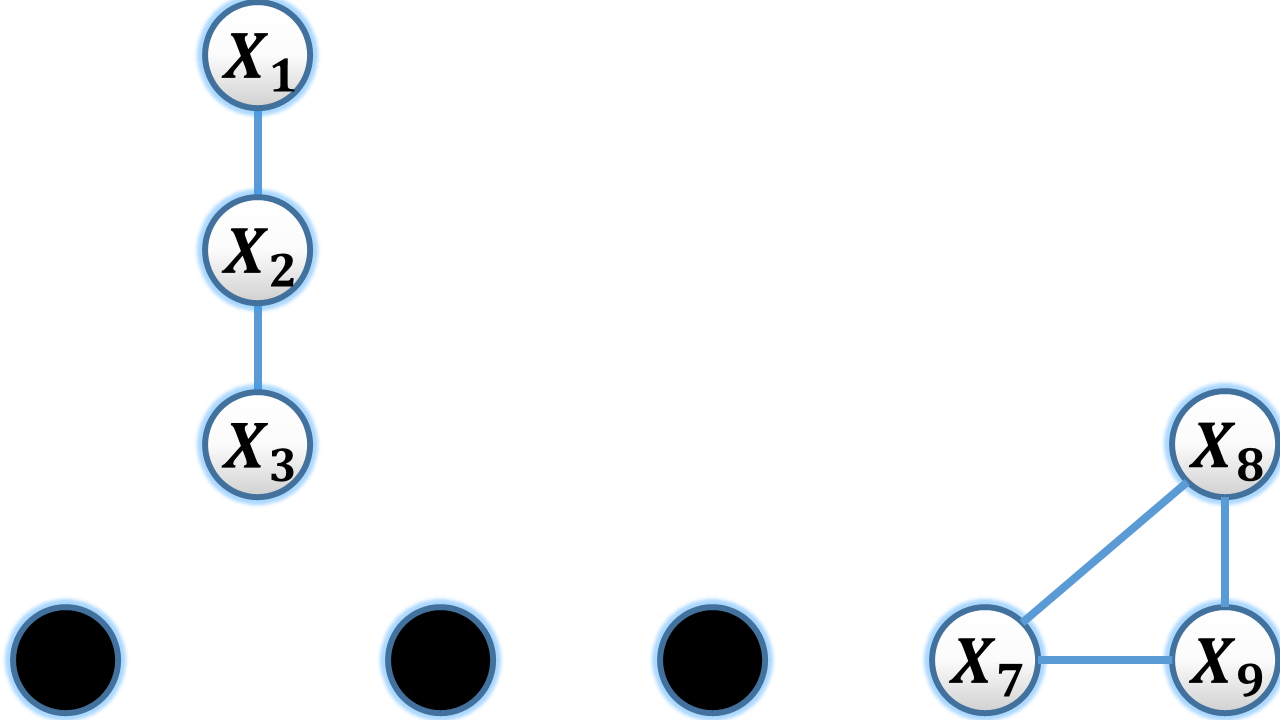}}
		\caption{$G_2$}
	\end{subfigure}
	\caption{(a) Graph $G_1$ is obtained from graph $G_0$ by removing directed edges. (b) After intervening on variable $X_4$, its incident edges will be oriented according to the true graph. (c) We apply meek rules to orient further edges.  (d) $G_2$ is obtained by removing directed edges.}
	\label{fig:step}
\end{figure}

\begin{example} An example of $G^*$ and its corresponding CPDAG are given in \figurename~\ref{fig:example}. In \figurename~\ref{fig:step},
 we illustrate one step of active causal structure learning  for this CPDAG. $G_1$ is obtained by deleting directed edges from $G_0$. At the first step, we choose variable $X_4$ for intervention. By performing perfect randomized experiment, the direction of incident edges to  $X_4$ are recovered which are $(X_3,X_4)$ and $(X_4,X_5)$. Next, we apply Meek rules and the orientation of $(X_3,X_5)$ will be discovered. We delete the oriented edges $(X_3,X_4)$, $(X_4,X_5)$ and $(X_3,X_5)$ and the graph $G_2$ is obtained from $G_1$ by deleting the directed edges. 
 \end{example}

\begin{figure*}[t!]
	\centering
	\includegraphics[width=\linewidth]{{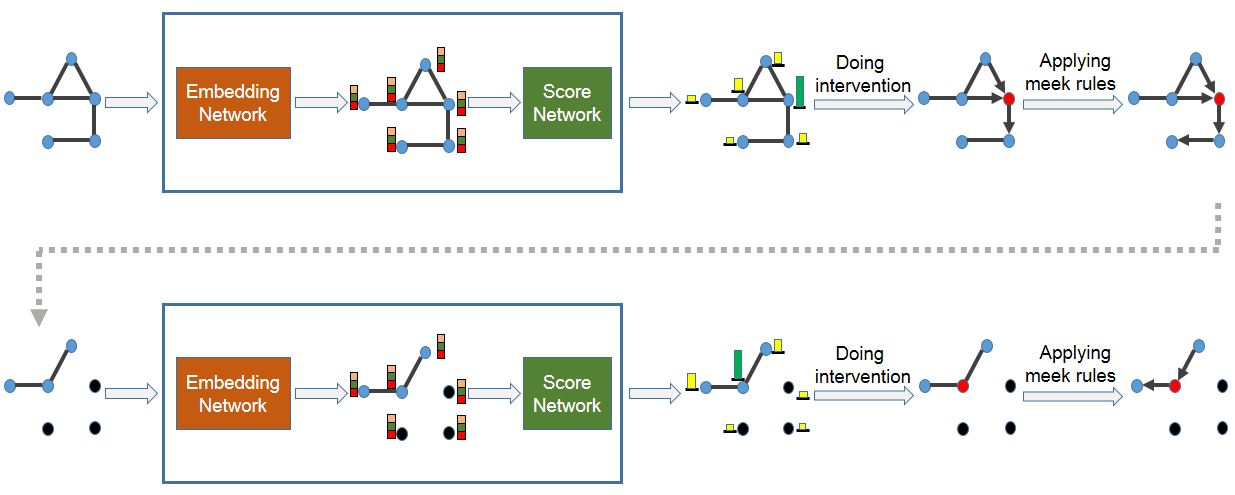}}
	\caption{An example of whole method for a given CPDAG. At first, the embedding vector of each node will be calculated by the ``embedding network''. Then the ``score network'' returns a score to each node based on its embedding vector. Afterwards, we select a node with the highest score for performing intervention. Next, some of the edges will be oriented as the result of intervention. We remove the directed edges and feed the remaining graph to ``Embedding network" for the next step.  }
	\label{fig:model}
\end{figure*}

\section{Proposed Method}

The proposed method is based on a greedy approach where for a given graph $G_j$ in each step $j$, we select a variable $X_{i_j}$ from graph $G_j$ based on a heuristic function $Q$. We train an agent by a reinforcement learning algorithm in order to obtain the function $Q$. More specifically, the function $Q$ has two arguments, namely, a chain graph $G$ and a variable $X$, and returns a score $Q(G,X)$ which represents how much it is desirable to select variable $X$ for doing intervention in graph $G$. Based on these scores, we choose the following variable in step $j$:
\begin{equation*}
X_{i_j} = \argmax_{X\in V(G_j)}Q(G_j,X).
\end{equation*}

Unlike previous works on experiment design in the active  setting, we will utilize deep reinforcement learning methods in order to find suitable $Q$ function. 
A diagram of proposed method for selecting a variable to be intervened on is given in \figurename~\ref{fig:model}. First, the embedding vector of each variable is calculated by ``embedding network". Then ``score network" determines score of each variable according to embedding vectors. The variable with the highest score will be selected for doing intervention. Direction of some edges will be discovered after performing intervention. Finally, the resulted graph is given as input for the next step and this process will be repeated until the whole causal structure is recovered.

\subsection{Embedding Network}
In order to feed the graph to the score network, we first need to represent the graph with a vector. Here, we use graph neural networks (GNN) \cite{xu2018powerful} to embed our graphs into vectors.

The GNNs use aggregated information of a variable's neighborhood to represent its embedded vector. To do so, the representation of a variable is updated by aggregating representations of its neighbors in an iterative manner. After $L$ iterations, a variable's representation captures the structural
information within its $L$-hop network neighborhood. We can also embed a whole graph by a pooling method, for instance, by summing the embedded vectors of all variables in the graph. In general, the $l$-th layer of a GNN can be written as follows:\\
\begin{equation}
\begin{split}
	a_v^{(l)}& = AGGREGATE^{(l)}(\{h_u^{(l-1)} : u \in N(v) \})\\
	h_v^{(l)}&=COMBINE^{(l)}(W_v,h_v^{(l-1)},a_v^{(l)}),
\end{split}
\end{equation}
where $h_v^{(l)} $ is the embedding vector of node $v$ at the $l$-th iteration/layer,
$N(v)$ is the set of nodes being adjacent to $v$, and $W_v$ is node features with a dimension of $q$. In our implementation, we set $W_v$ to the vector of all ones.
Moreover, the $h_v^{(0)}$ is initialized by $W_v$. Multiple options have been proposed for operations  $AGGREGATE^{(l)}(.)$ and $COMBINE^{(l)}(.)$  in the literature of GNNs.
Here, we use $SUM$ function for $AGGREGATE$, and $RELU$ for $COMBINE$. Hence, we can rewrite the above equation as follows:

\begin{equation}
\begin{split}
h_v^{(l)} = RELU\Big(\theta_1 W_v + \theta_2 \sum_{u \in N(v)}{h_u^{(l-1)}}  \Big), \mbox{for $l=1,\cdots,L$},
\end{split}
\end{equation}
where $\theta_1 \in R^{p\times q}$ and $\theta_2 \in R^{p\times p}$ are the parameters of the model.

\subsection{Score Network}
In the literature of causality, multiple heuristic functions have been proposed for the problem of active causal structure learning which are mainly based on computing the size of MEC \cite{andersson1997characterization}. However, here, we parameterized the heuristic function and denote it by $\hat{Q}$ where its parameters are needed to be trained. More specifically, we consider the following parameterized heuristic function:\\
\begin{equation}
\hat{Q}(G_i,v;\Theta)=\theta_3^T RELU([\theta_4\sum_{u\in V}{h^{(L)}_u},\theta_5 h^{(L)}_{v}]),
\end{equation}
where $h^{(L)}_{v}$ is $p$-dimension embedding vector of node $v$ after $L$ iterations, $\theta_3 \in R^{2p}$, $\theta_4 ,\theta_5 \in R^{p\times p}$ and $[.,.]$ is the concatenation operator. We denote the set of all parameters by $\Theta=\{\theta_i\}_{i=1}^5$.

\subsection{Training Phase}

There is an analogy between selecting a variable for intervention and taking actions by an agent in an unknown environment. In particular, the state-action-value function defined in reinforcement learning problem determines the overall expected reward of doing each action in each state. In our problem, we can use similar function to determine which variable is more desirable for performing intervention in each step. 

\subsubsection{Reinforcement Learning Formulation}
In the following, we explain how our problem can be formulated in the framework of reinforcement learning by introducing the set of states, the set of actions, and the reward function: 
\begin{itemize}
	\item Set of states: We consider all embedding vectors of chain graphs as the set of states. Thus, at each time step $j$, the embedding vector of $G_j$ represents the state of the system. 
	\item Set of actions: We consider intervening on any variable in the system as the set of actions.
	\item Reward function: For a given graph $G$ and a variable $X$ in the system, we consider the reward function as the number of directed edges that can be oriented after intervening on $X$ in the chain graph $G$. 
\end{itemize}
Our goal is to find an optimal policy that maximizes the expectation of overall reward which is the total number of oriented edges.

\begin{algorithm}[t]
	\caption{Q-learning algorithm}
	\begin{algorithmic}
		\STATE Initialize $\Theta$ and set $\epsilon=1$ and experience memory $M=\emptyset$ for the replay buffer
		\FOR{episode $e = 1$ to $E$}
		\STATE Sample a DAG $G^*$
		\STATE Create CPDAG $G_1$
		\STATE Initialize  $I = \{\} $
		\FOR{step $j=1$ to $T$}
		\STATE 
		
		$X_{i_j} = 
		\begin{cases}
		\text{random node } X \in V(G_j) & \text{w.p. } \epsilon(e) \\
		\displaystyle \argmax_{X\in V(G_j)}\hat{Q}(G_j,X;\Theta) & \text{o.w}
		\end{cases}
		$
		
		\STATE Add $X_{i_j}$ to $I$
		\STATE Direct edges that are connected to $X_{i_j}$ according to $G^*$ and applying Meek rules
		\STATE Remove directed edges to obtain $G_{j+1}$
		\STATE Add tuple $(G_j,X_{i_j},r_j,G_{j+1})$ to $M$
		\ENDFOR
		\IF {$e$ mod $q=0$}
		\STATE Sample random batch $B$ from $M$
		\STATE Update $\Theta$ by $B$
		\ENDIF
		\STATE Decrease $\epsilon(e)$
		\ENDFOR
	\end{algorithmic}
\end{algorithm}

\newcommand{\ourMethodName}{Proposed}

\begin{table*}[!t]
	\caption{Running times of algorithms for performing five interventions (in seconds), $\rho=0.1,0.2,0.3$. The values in the parentheses show the speedup factor of our proposed method with respect to the considered algorithm.}
	\label{tab:time}
	\centering
	\begin{tabular}{|l|l|l|l|l|l|l|l|l|l|}
		\hline
		\multicolumn{2}{|l|}{Nodes}   & 15    & 20    & 25    & 30    & 35   & 40    & 50    & 70    \\ \hline
		\multirow{3}{*}{\ourMethodName}& $\rho=0.1$ & 0.001 & 0.002 & 0.003 & 0.005 & 0.011 & 0.021  & 0.16  & 0.91  \\\cline{2-10}
		& $\rho=0.2$ & 0.002 & 0.004 & 0.007 & 0.012 & 0.03 & 0.09  & 0.25  & 1.6  \\\cline{2-10}
		& $\rho=0.3$ & 0.050 & 0.069 & 0.120 & 0.136 & 0.198 & 0.562  & 0.623  & 3.7  \\\hline
		\multirow{3}{*}{Average}& $\rho=0.1$ & 0.03 (30)  & 0.074 (35)  & 0.17 (57)  & 0.38 (76)  & 1.01 (92) & 2.02 (96) & 14.4 (90) & 99 (109)  \\\cline{2-10}
		& $\rho=0.2$ & 0.08 (40)  & 0.25 (62)  & 0.65 (93)  & 0.93 (77)  & 4.52 (150) & 17.71 (197) & 41.37 (165) & 220 (137)  \\\cline{2-10}
		& $\rho=0.3$ & 0.092 (2)  & 0.311 (4)  & 0.841 (7)  & 2.68 (20)  & 13.00 (66) & 34.0 (60) & 472 (757) & 607 (164) \\\hline
		\multirow{3}{*}{Minimax}& $\rho=0.1$ & 0.73 (730)  & 2.43 (1215)    & 7.34 (2446)   & 35.7 (7140)  & 381 (34636) & 1811 (86238) &   -   &  -    \\\cline{2-10}
		& $\rho=0.2$ & 3.23 (1615)  & 22.4 (5600)   & 185.3 (26471)  & 1479 (123250) & - &  -    &   -   &  -     \\\cline{2-10}
		& $\rho=0.3$ & 18.15 (353) & 254.73  (3691)  & 1193 (9941)  &  -  &  -   &   -   &   -   & -     \\\hline
		\multirow{3}{*}{Entropy}& $\rho=0.1$ & 0.21 (210)  & 0.71 (355)    & 2.86 (953)   & 11.2 (2240)  &   222 (20182)   & 1409 (67095)      &  -    &  -    \\\cline{2-10}
		& $\rho=0.2$ & 4.56 (2280) & 31 (7750)   & 304 (43428)  & 7279 (606583) &   -  &   -   &   -   &  -   \\\cline{2-10}
		& $\rho=0.3$ & 27.31 (546) & 382.57 (5544)  & 4998 (41650)  & -  &   -  &  -    &   -   &  -   \\\hline	
	\end{tabular}
	\vspace{ - 05 mm}
\end{table*}

\subsubsection{Learning Q Function}
We utilize $Q$-learning method in deep reinforcement learning (DRL) \cite{sutton2018reinforcement} to obtain $\hat{Q}$ function which is an iterative algorithm that updates the value of $Q$ function in each step. In DRL, $Q$ function can be presented by a look-up table where the value of $E(\sum_{t=1}^{T}\gamma^t r_t|G,X)$ is given for each pair of $(G,X)$ where $r_t$ is the reward in step $t$, $\gamma$  is a discount factor in the range $(0,1)$ and $T$ is number of steps in an episode. In many applications, the state space is so huge that we cannot observe all states in the training process or even keep them in the look-up table. To resolve this issue, we obtain an approximation of it by training the parameters of score network through $Q$-iteration algorithm. To do so, we consider two steps in each iteration: updating state-action-value function and updating the weights of networks.

The Q-learning \cite{sutton2018reinforcement,riedmiller2005neural} updates can be written as follows:
\begin{align}
	&\hat{Q}(G_j,X_{i_j};\Theta) \leftarrow \hat{Q}(G_j,X_{i_j};\Theta)+\nonumber\\
	& \alpha \Big(r(G_{j}) + \gamma \max_{X \in V(G_{j+1})}{\hat{Q}(G_{j+1},X;\Theta)} ~~- \hat{Q}(G_j,X_{i_j};\Theta)\Big)
\end{align}
where $r(G_{j})$ is the number of edges that are oriented as a result of intervening on $X_{i_j}$ in graph $G_j$ and $\alpha$ is the learning factor. For updating $\Theta$, we use gradient descent method:
\begin{align}
	& \Theta \leftarrow \Theta + 
	 \nabla_{\Theta} \Bigg( \big(r(G_{j+1})+\nonumber\\& \gamma \max_{X \in V(G_{j+1})}{\hat{Q}(G_{j+1},X;\Theta)} - \hat{Q}(G_j,X_{i_j};\Theta) \big)^2 \Bigg).
\end{align}

The description of Q-learning algorithm is given in Algorithm 1. At the beginning, we initialize $\Theta$ randomly according to normal distribution $N(0,1)$ and consider an empty replay buffer. 
Then we start outer loop where in each iteration, we sample a DAG $G^*$, and construct CPDAG $G_1$ from $G^*$. We initialize intervention sets $I$ to $\emptyset$. In each iteration of inner loop, we select a variable to intervene on based on $\epsilon$-greedy algorithm in order to explore more states in earlier episodes. 
We orient incident edges of the selected node based on the true causal graph, and then apply meek rules. Next, we remove oriented edges from the graph $G_j$ and obtain $G_{j+1}$. We add tuple $(G_j,X_{i_j},r_j,G_{j+1})$ to the replay buffer $M$. The current episode is finished once the inner loop is complete. At this point, we decrease $\epsilon$ for the next execution of the inner loop. After iterating $q$ number of episodes, we sample a batch $B$ from $M$ and update $\Theta$ accordingly.

 \newcommand{\resultFolder}{final_res_0_1}
 \begin{figure*}
 	\centering
 	\begin{subfigure}[t]{0.305\linewidth}
 		\includegraphics[width=\linewidth]{{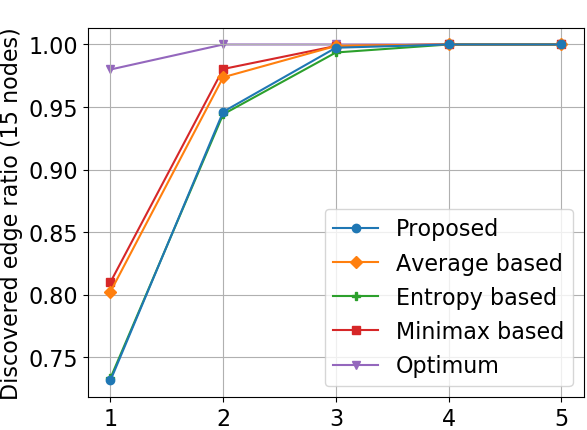}}		
 	\end{subfigure}
 	\begin{subfigure}[t]{0.29\linewidth}
 		\includegraphics[width=\linewidth]{{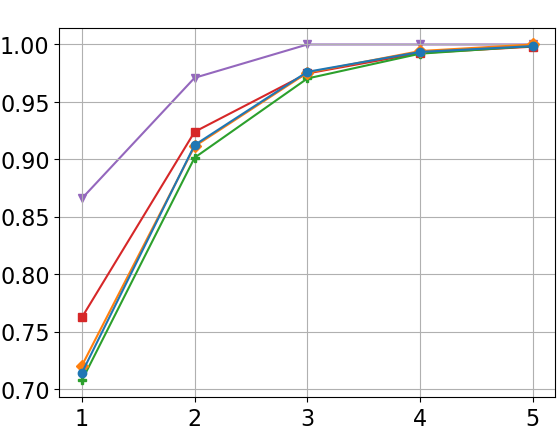}}
 	\end{subfigure}
 	\begin{subfigure}[t]{0.29\linewidth}
 		\includegraphics[width=\linewidth]{{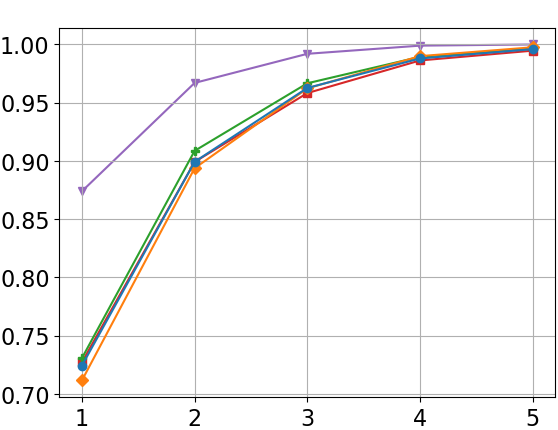}}	
 	\end{subfigure}
 	
 	\begin{subfigure}[t]{0.30\linewidth}
 		\includegraphics[width=\linewidth]{{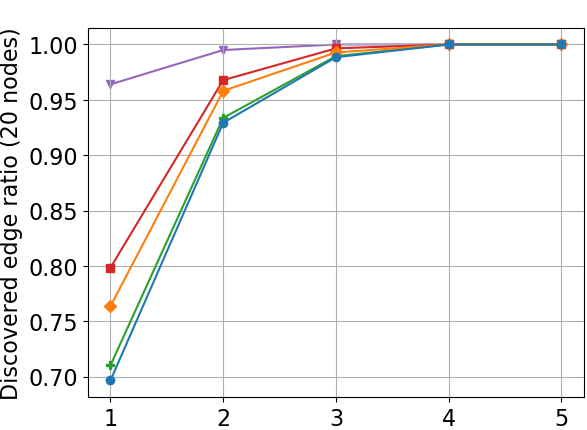}}
 	\end{subfigure}
 	\begin{subfigure}[t]{0.29\linewidth}
 		\includegraphics[width=\linewidth]{{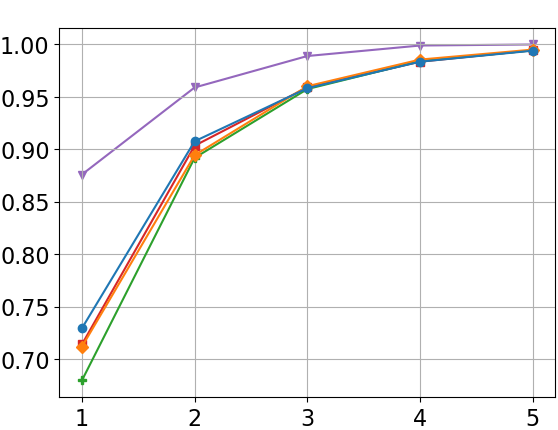}}
 	\end{subfigure}
 	\begin{subfigure}[t]{0.29\linewidth}
 		\includegraphics[width=\linewidth]{{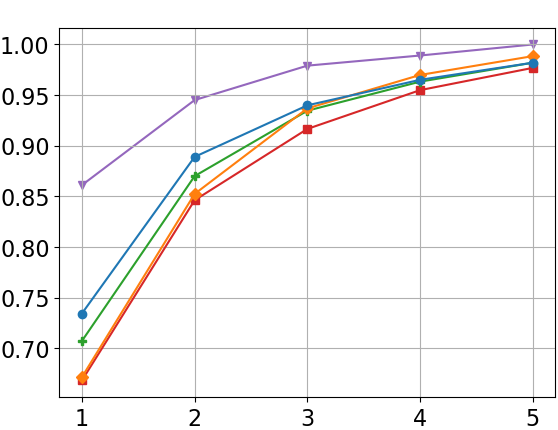}}	
 	\end{subfigure}
 	
 	\begin{subfigure}[t]{0.305\linewidth}
 		\includegraphics[width=\linewidth]{{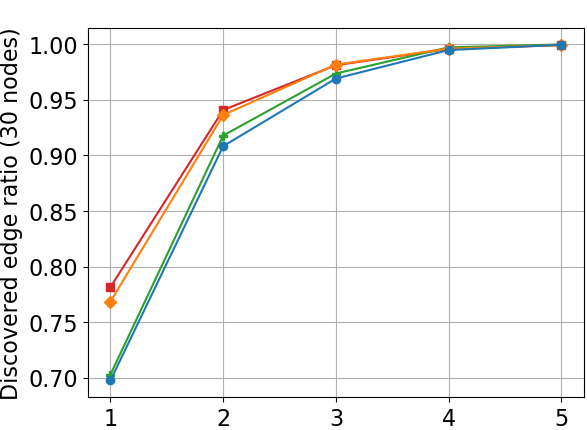}}
 	\end{subfigure}
 	\begin{subfigure}[t]{0.29\linewidth}
 		\includegraphics[width=\linewidth]{{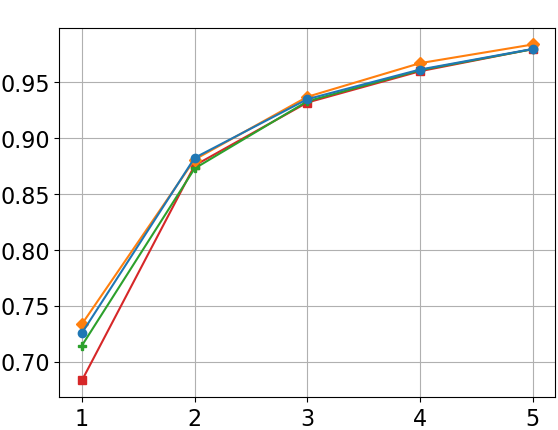}}
 	\end{subfigure}
 	\begin{subfigure}[t]{0.29\linewidth}
 		\includegraphics[width=\linewidth]{{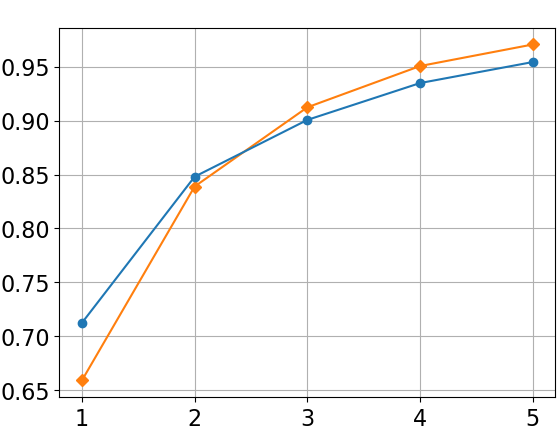}}
 	\end{subfigure}
 	
 	\begin{subfigure}[t]{0.305\linewidth}
 		\includegraphics[width=\linewidth]{{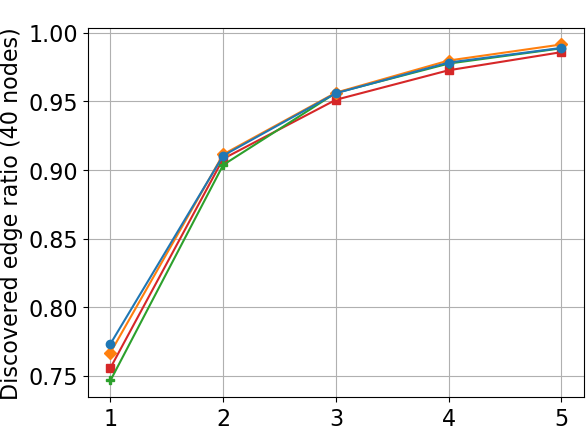}}
 	\end{subfigure}
 	\begin{subfigure}[t]{0.29\linewidth}
 		\includegraphics[width=\linewidth]{{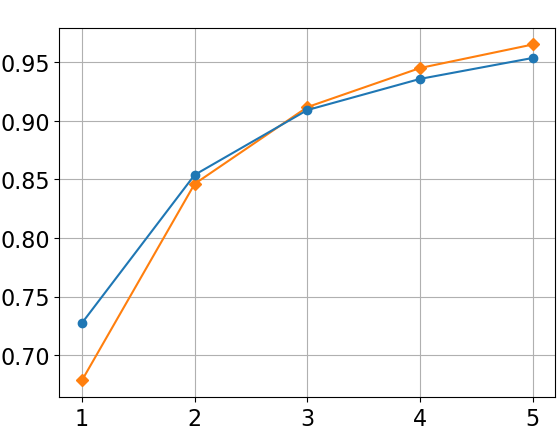}}
 	\end{subfigure}
 	\begin{subfigure}[t]{0.29\linewidth}
 		\includegraphics[width=\linewidth]{{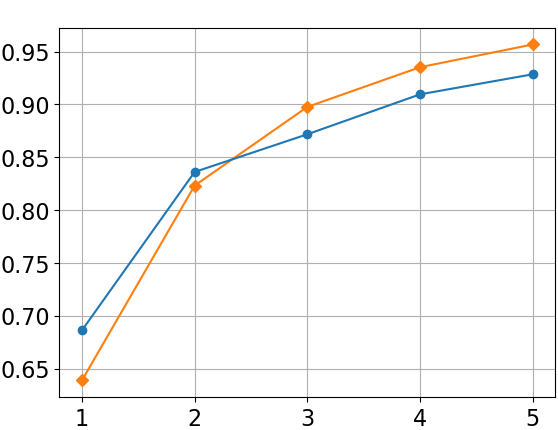}}
 	\end{subfigure}
 	
 	\begin{subfigure}[t]{0.305\linewidth}
 		\includegraphics[width=\linewidth]{{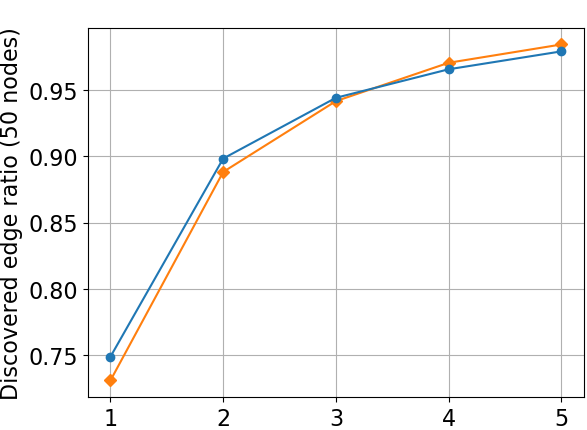}}
 	\end{subfigure}
 	\begin{subfigure}[t]{0.29\linewidth}
 		\includegraphics[width=\linewidth]{{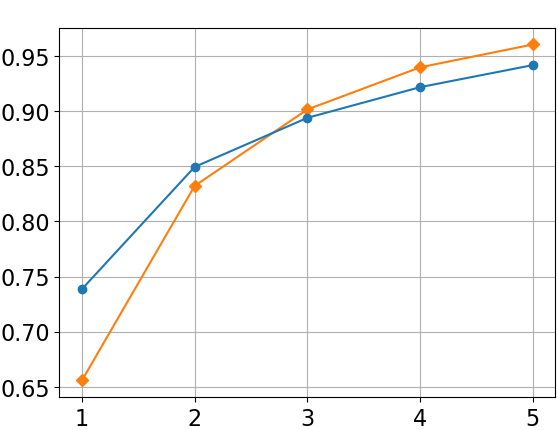}}
 	\end{subfigure}
 	\begin{subfigure}[t]{0.29\linewidth}
 		\includegraphics[width=\linewidth]{{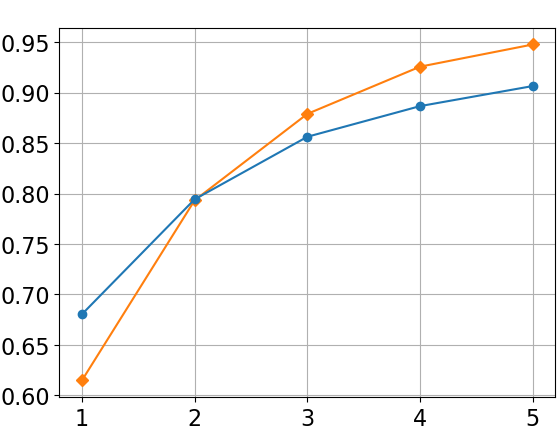}}
 	\end{subfigure}
 	
 	\begin{subfigure}[t]{0.305\linewidth}
 		\includegraphics[width=\linewidth]{{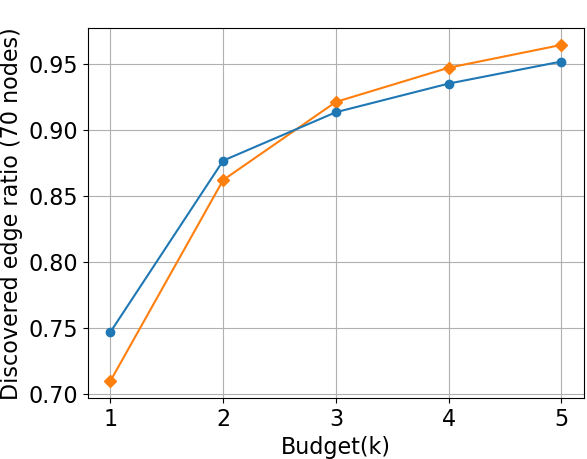}}
 	\end{subfigure}
 	\begin{subfigure}[t]{0.29\linewidth}
 		\includegraphics[width=\linewidth]{{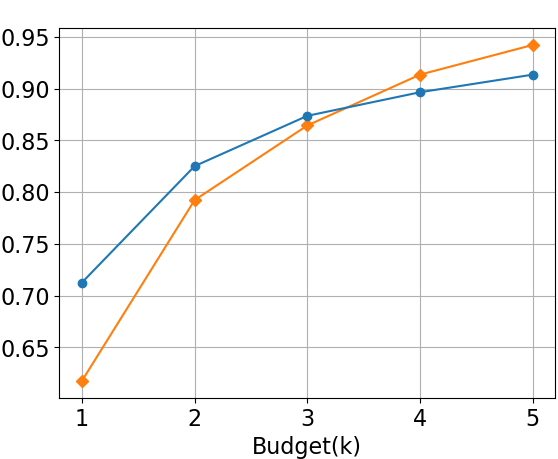}}
 	\end{subfigure}
 	\begin{subfigure}[t]{0.29\linewidth}
 		\includegraphics[width=\linewidth]{{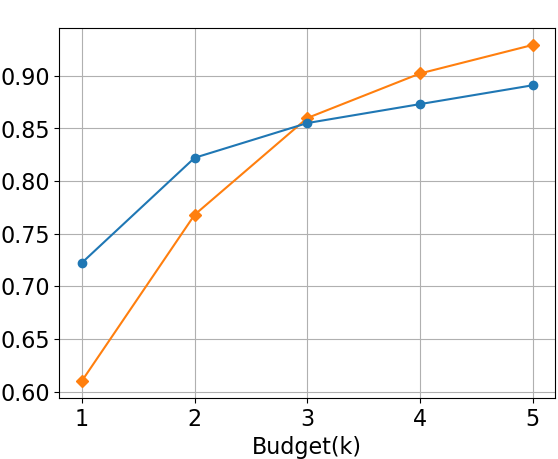}}
 	\end{subfigure}
 	
 	\caption{Performance of different algorithms on graphs with $\rho=0.1$ (left), $\rho=0.2$ (middle) and $\rho=0.3$ (right) in terms of discovered edge ratio. The first row to the sixth row correspond to $n=15,20,30,40,50,70$.
 	}
 	\label{fig:results}
 \end{figure*}

 \begin{figure*}[t]
	\centering
	\begin{subfigure}{0.305\linewidth}
		\includegraphics[width=\linewidth]{{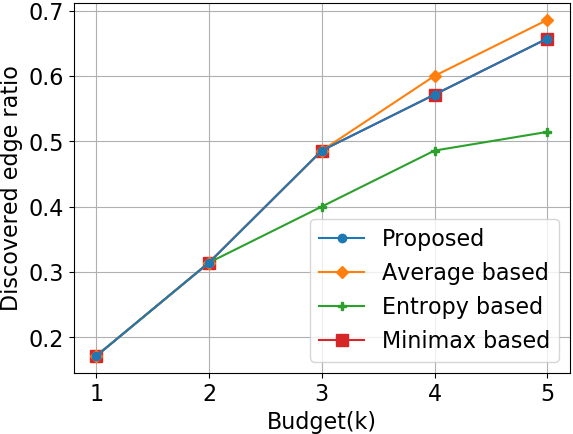}}	
		\caption{Yeast1}
	\end{subfigure}
	\begin{subfigure}{0.29\linewidth}
		\includegraphics[width=\linewidth]{{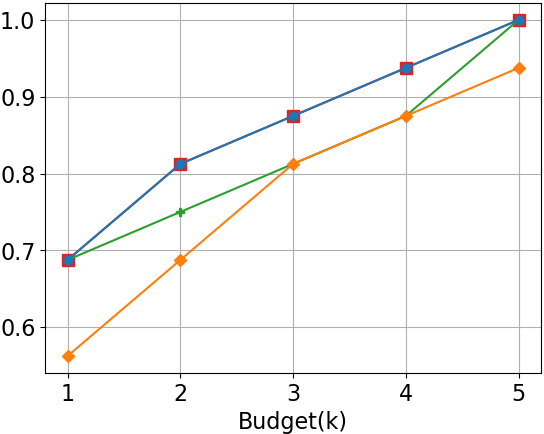}}
		\caption{Yeast2}
	\end{subfigure}
	\begin{subfigure}{0.29\linewidth}
		\includegraphics[width=\linewidth]{{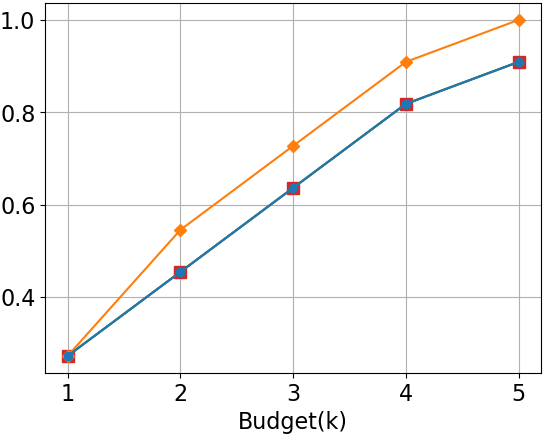}}
		\caption{Yeast3}
	\end{subfigure}
	
	\begin{subfigure}{0.305\linewidth}
		\includegraphics[width=\linewidth]{{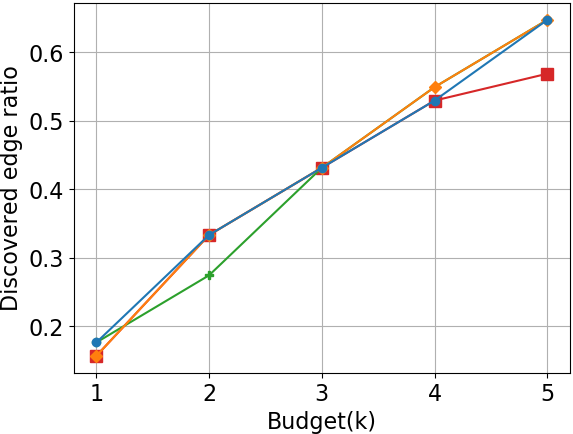}}		
		\caption{Ecoli1}
	\end{subfigure}
	\begin{subfigure}{0.29\linewidth}
		\includegraphics[width=\linewidth]{{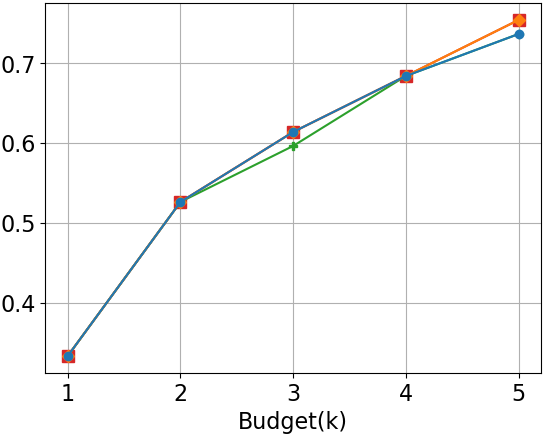}}
		\caption{Ecoli2}	
	\end{subfigure}
	\caption{Comparing performance of the proposed algorithm against previous methods in real graphs.}
	\label{fig:realData}
\end{figure*}

\section{Experimental Results}

\subsection{Synthetic Graph Generation}

We employ synthetically generated chordal graphs similar to the ones considered in \cite{ghassami2017budgeted} in order to evaluate the performance of different algorithms. To do so, a randomly chosen perfect elimination ordering (PEO) \cite{rose1978algorithmic} over the vertices is used to generate the underlying chordal graphs. Starting from the vertex $v$ with the highest order, all the vertices  with lower orders are connected to $v$ with probability inversely proportional to the order of $v$. Then, all the parents of $v$ are connected with directed edges, where each edge is directed from the parent with the lower order to the parent with the higher order. If vertex $v$ is not connected to any of the vertices with the lower order, one of them is taken uniformly at random and set as the parent of $v$. In this way, we make sure that the generated graph will be connected.

\subsection{Previous Works for Quantitative Comparisons}
We consider the following three main related works that have been proposed previously for the active setting:

\begin{itemize}
	\item Entropy based approach \cite{he2008active}:
	A heuristic function based on Shannon's entropy metric is used such that the MEC can be reduced by intervening on the selected variable into a subclass as small as possible.
	\item Minimax based approach \cite{hauser2012characterization}: Huaser and Buhlmann proposed an optimal algorithm for performing a single intervention according to a minimax objective function. This algorithm  is utilized as an heuristic function for selecting a variable for intervention in each step.
	\item Average based approach \cite{ghassami2017budgeted}:
	A variable is selected for intervention in each step which maximizes the expected number of edges whose orientations can be recovered by performing  intervention on this variable.
\end{itemize}

\subsection{Details of Training and Testing Phases:}

The experimental results are obtained from a model that was trained by graphs with the number of nodes in the set $\{20,25\}$. All the hyper-parameters were selected based on the results for these small graphs and then used for all the other graphs. For the testing phase, we generated $50$ instances of chordal graphs with $\{15,20,25,30,35,40,50,70\}$ nodes and density $\rho=0.1,0.2,0.3$ which is the average number of edges divided by ${n \choose 2}$. The implementation of our method is available in the supplementary material.

All procedures are executed on a Xeon server with $16$ cores operating at $2.10$~GHz. 
To train the proposed model, we use multi-thread programming. However, the test mode uses only a single thread, i.e., is not multi-threaded. 
For the related works, we implemented average based and minimax based approaches by multi-thread programming in order to improve their running times.

\subsection{Comparison Results}

To compare the performance of the above algorithms, we considered discovered edge ratio as the performance measure which is the ratio of oriented edges after performing experiments to the total number of edges in the graph.  In our experiments, we assumes that the budget of interventions is equal to $5$.

In \figurename~\ref{fig:results}, the average discovered edge ratio is plotted against the number of interventions for the considered algorithms. We reported the performance of an algorithm if it returns the output within at most three hours. Different rows show the results for different number of variables. The left, middle, and right charts in every row show the results for graphs with density $\rho=0.1$, $0.2$, and $0.3$, respectively. 
In the first two rows, we also depict the results of optimal solution. 
Comparing with the other approaches, for graphs with high density, our method always has better performance for $k=1,2$. For graphs with low density, our method has a lower performance. However, by increasing the graph size, the performance of our method gets close the ones of other approaches. 

We also compare the algorithms in terms of their running times. We measure running times for five number of interventions on each graph. In table \ref{tab:time}, the average running times for different graphs are provided. Moreover, in every entry, we also provide the speedup factor of our proposed method with respect to the considered algorithm in parentheses. As can be seen, running times of entropy based approach and minimax based approach grow exponentially with the graph size. We did not report running times of an algorithm if it did not terminate after three hours. 
Compared to the second best solution (i.e., average based approach), our method reduces running time by a factor of up to $757$ in dense graphs.

According to the results, we can conclude that the proposed method generalizes fairly well to graphs with different sizes although it has been trained over a specific range of graph sizes. For experiment design with limited budget of $k\leq 3$, the proposed method has the best performance compared to other related works in most cases. Finally, it significantly reduces running times with respect to existing solutions.

\subsection{Real Graphs}

In addition to synthetic graphs, we also did experiment with Gene Regulatory Networks (GRN). GRN is a network of biological regulators that interact with each other. In GRN, there exist some transcription factors which have direct impacts on gene activations. More specifically, the interactions between transcription factors and regulated genes can be shown by a directed graph where there is a direct edge from a transcription factor to a gene if it regulates the gene expression.

We consider the GRN in ``DREAM 3 In Silico Network" challenge \cite{marbach2009generating}. The networks in this challenge were extracted from known biological interactions in GRN of E-coli and Yeast bacteria. The size of each sub-network is equal to 100. For each sub-network, we obtain CPDAG from the true causal network and provide it as an input to the algorithms. Fig.~\ref{fig:realData} illustrates the discovered edge ratio of the algorithms in five real sub-networks. As can be seen, the proposed method achieves competitive performance in most sub-networks.

\section{Conclusion}

In this paper, we proposed a deep reinforcement learning based solution for the problem experiment design. In the proposed solution, we embed input graphs to vectors using a graph neural network and feed them to another neural network which gives scores to variables in order to select the intervention target in the next step. We jointly train both neural networks by Q-iteration algorithm. Experimental results showed that the proposed solution has competitive performance in recovering the causal structure with respect to previous works which are mainly based on heuristic metrics related to graph properties of MEC. Moreover, the proposed solution reduces running times significantly and can be applied on large graphs.

\bibliographystyle{IEEEtran}
\bibliography{DRLGraph}

\end{document}